\documentclass[a4paper,11pt]{article}

\usepackage[pdftex]{graphicx}

\usepackage{amsmath}
\usepackage{hyperref}
\usepackage{amssymb}

\usepackage{geometry}
\usepackage{color}

\begin{document}

\author{Xikun~Zhang, Chang~Xu, Xinmei~Tian, and~Dacheng~Tao}
\title{Graph Edge Convolutional Neural Networks for Skeleton Based Action Recognition}
\maketitle

\begin{abstract}
This paper investigates body bones from skeleton data for skeleton based action recognition. Body joints, as the direct result of mature pose estimation technologies, are always the key concerns of traditional action recognition methods. However, instead of joints, we humans naturally identify how the human body moves according to shapes, lengths and places of bones, which are more obvious and stable for observation. 
Hence given graphs generated from skeleton data, we propose to develop convolutions over graph edges that correspond to bones in human skeleton. We describe an edge by integrating its spatial neighboring edges to explore the cooperation between different bones, as well as its temporal neighboring edges to address the consistency of movements in an action. A graph edge convolutional neural network is then designed for skeleton based action recognition. Considering the complementarity between graph node convolution and graph edge convolution, we additionally construct two hybrid neural networks to combine graph node convolutional neural network and graph edge convolutional neural network using shared intermediate layers. Experimental results on Kinetics and NTU-RGB+D datasets demonstrate that our graph edge convolution is effective to capture characteristic of actions and our graph edge convolutional neural network significantly outperforms existing state-of-art skeleton based action recognition methods. Additionally, more performance improvements can be achieved by the hybrid networks. 
\end{abstract}


\section{Introduction}

Human action recognition is one of the most challenging and important computer vision tasks. It has received considerable attention from both academia and industry, due to its wide application in video surveillance, virtual reality, human-computer interaction and robotics. Classical studies used to investigate action recognition based on monocular RGB videos \cite{wang2007learning,simonyan2014two,wang2011action,wang2013action,le2011learning,liu2018computational}, which is difficult to comprehensively represent actions in 3D space. Given the fast development of low-cost devices to capture 3D data (e.g. camera arrays and Kinect), increasingly more researches are actively conducted over 3D action recognition \cite{vieira2012stop,chen2016real,luo2013group,wang2012robust,yang2012eigenjoints}.

Skeletons generated from depth maps are often invariant to viewpoint or appearance. Most importantly, as a high-level abstraction of human actions, skeleton data greatly simplifies the difficulty in representing and understanding different action categories. Recently, different technologies have been developed to estimate the temporal motion of the skeleton by tracking and analyzing the motion of human joints. One of the pretty straight-forward approaches is to concatenate coordinates of joints into a long feature vector at each time step, and then feed it into temporal analysis models \cite{wang2012mining,fernando2015modeling}. However, spatial relationship between joints, as an indispensable part of human skeleton, has been neglected. To exploit connections between joints, \cite{hussein2013human} used covariance matrix for skeleton joint locations over time as a discriminative descriptor; \cite{wang2012mining} proposed to represent an action by a weighted summation of actionlets. To automatically capture the patterns embedded in spatial configuration and temporal dynamics of joints, \cite{yan2018spatial} proposed spatial-temporal graph convolutional networks to conduct convolution on human skeleton in both spatial and temporal domain \cite{li2015gated,scarselli2009graph,bruna2013spectral,henaff2015deep,niepert2016learning}. \cite{li2017skeleton} proposed a two-stream CNN to process both raw coordinates and motion data obtained by subtracting joint coordinates in consecutive frames; \cite{ke2017new} designed a model that first transforms skeleton sequence into three clips corresponding to three cylindrical coordinates of skeleton sequence, and then applies deep CNN. %\cite{soo2017interpretable} proposed temporal convolutional neural networks applied to 3D human action recognition that provides us a way to explicitly learn readily interpretable spatio-temporal representations for 3D human action recognition.

These methods have brought in impressive performance improvements of action recognition. They mostly focus on human joints, since movements originate at joints. But it is also instructive to note the fact that a joint connects two bones, whose shapes, lengths and places dictate how the human body moves in practice. In most cases, some joints without obvious changes of coordinates in 3D space would still be able to drive the significant motion of bones, e.g. the hip and shoulder joints during walking or running. On the other hand, when we humans identify the actions of a person, we may ignore movements of particular joints, but we must always note the obvious motions of bones. Hence, instead of calculating subtle changes of body joints, movements magnified on bones deserve more attention.

In this paper, we revisit skeletal data from the perspective of bones, and propose an edge convolutional neural network for action recognition. Different from conventional graph convolutional neural networks concentrating on graph nodes \cite{ke2017new,soo2017interpretable,yan2018spatial}, we focus on invaluable information carried by graph edges (e.g. bones generated from skeleton data), while graph nodes only indicate connections between edges. 
We develop graph edge convolution, which represents an edge by learning weights to integrate its spatial neighboring edges (see Fig. 1). To handle temporal changes of skeleton data in continuous video frames, we extend graph edge convolution by including temporal neighboring edges into consideration. In addition, graph node and edge convolutions process skeleton data from two different perspectives, and they are complementary with each other. 
We design two different hybrid networks to integrate results of these two types of graph convolutions, one is through a shared dense layer and the other is to add two common convolutional layers. We evaluate the proposed graph edge CNN on two datasets Kinetics and NTU-RGB+D, which suggests the advantages of the proposed graph edge convolution over existing state-of-the-art techniques, and the combination of graph edge and node convolutions can lead to further performance improvement.

The organization of this paper is as follows. In the following section, we review some related works, and in Section \uppercase\expandafter{\romannumeral3}, we introduce our graph edge convolutional neural networks as well as two hybrid models integrating node and edge convolution. In Section \uppercase\expandafter{\romannumeral4}, the experimental results are presented. Finally, we draw some conclusions in Section \uppercase\expandafter{\romannumeral5}.

\begin{figure}
\centering
\includegraphics[height=3.8cm]{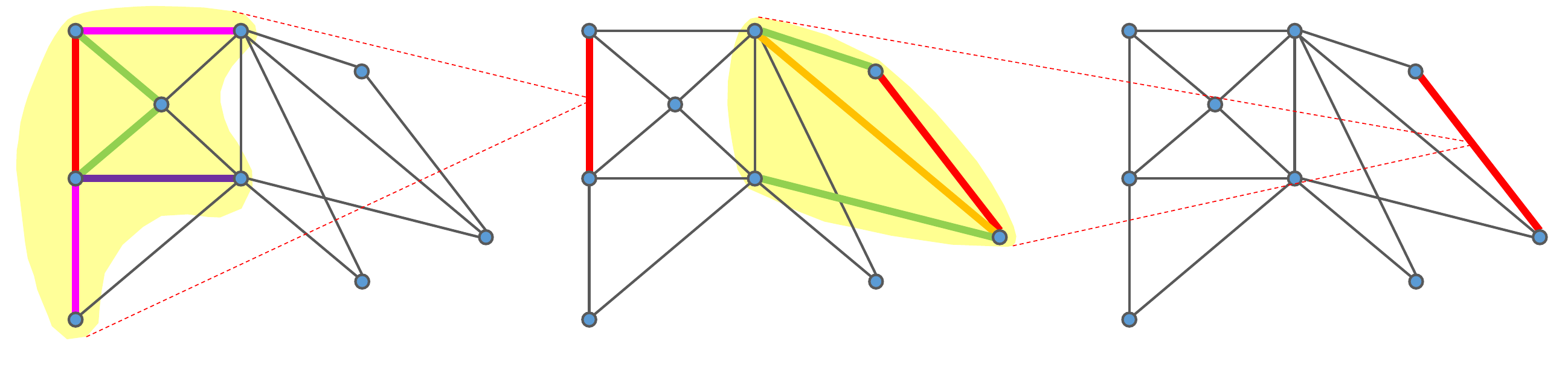}
\caption{Illustration of edge convolution operation. In {\it left} and {\it middle graph}, the {\it light yellow zone} are receptive fields, the {\it red bold edge} is the center edge in receptive field. Other {\it colored bold edges} besides the {center red edge} are neighboring edges, and their difference in color indicates different weights assigned to them.}
\label{fig:example}
\end{figure}

\section{Related Work}
In this section, we will briefly review related works on skeleton based action recognition and graph convolutional neural networks.
\subsection{Graph Convolutional Networks}
Generalizing convolutional networks from images to graph structured data has been an active topic in recent years. \cite{bruna2013spectral,henaff2015deep,li2017skeleton,niepert2016learning}. These works mainly fall into two categories: 1) Spectral perspective, which converts graph data into its spectrum and apply CNNs on spectral domain, for example, \cite{bruna2013spectral} defines a spectral convolutional layer to apply filters on frequency domain; \cite{henaff2015deep} develops an extension of spectral networks which incorporates a graph estimation procedure, and applies their model on ImageNet dataset. There are many similar works in this stream \cite{duvenaud2015convolutional,li2015gated,kipf2016semi}, but a fundamental limitation of the spectral method has to be stressed is that the spectral construction is limited to a single domain. The reason is that the spectral filter coefficients are basis dependent. If we learn a filter w.r.t to a certain set of basis, this filter will not be able to be applied to another domain with another set of basis. This problem can be solved if we can construct compatible orthogonal bases across different domains. However, such construction requires the knowledge of some correspondence between domains, which is extremely hard in most cases. For example, in computer graphics applications, finding correspondence between shapes is itself a very hard problem; 2) Spatial perspective, works in this stream directly design convolution operation on spatial domain, which resembles the convolution on images a lot \cite{li2015gated,scarselli2009graph}. Applying graph CNN designed on spatial domain to solve problems in computer vision is also attracting more and more attention: \cite{such2017robust} utilize polynomials of functions of graph adjacency matrix as filters to conduct graph convolution, then apply it to image classification and 3D mesh classification; \cite{lai2017learning} designs depthwise separable graph convolution and apply it to do image classification on CIFAR datasets; \cite{simonovsky2017dynamic} formulates a convolution-like operation on graph signals performed in the spatial domain where filter weights are conditioned on edge labels and dynamically generated for each specific input sample. Our work also proposes a new graph convolution model, and we apply it to human action recognition.

\subsection{Skeleton Based Human Action Recognition}
The development of highly accurate depth sensors and pose estimation algorithms \cite{shotton2011real,cao2017realtime} makes it easy to get skeletal data of human action, and skeleton based human action recognition is attracting more and more interests. The work on this topic mainly follows two streams: {\bf 1) Hand-crafted features}. Work belonging to this stream leverages the dynamics of joint motion by handcrafted features. For example, \cite{hussein2013human} used covariance matrix for skeleton joint locations over time as a discriminative descriptor for a sequence, and used multiple covariance matrices to encode the temporal dependency of joint locations; \cite{wang2012mining} used actionlet ensemble obtained by data mining to represent action, and designed LOP feature to overcome intra-class variance caused by the imperfectness of raw data; \cite{xia2012view} utilized histograms of 3D joint locations (HOJ3D) as a compact representation of postures, then reprojected HOJ3D using LDA and clustered it into k posture visual words; \cite{vemulapalli2014human} proposed to explicitly model 3D geometric relationships between various body parts using rotations and translations in 3D space to represent human skeleton, and then model human actions as curves in a Lie group; \cite{kim2014saliency} describes a new 3D saliency prediction model, in which salient 3D space-time regions in a video are detected and segmented, and then the saliency strength of each segment is calculated using different attributes including motion, disparity, texture, and the predicted degree of visual discomfort experienced. {\bf 2) Deep learning approaches}, which processes the skeletal data by deep learning methods. \cite{du2015hierarchical} divides human skeleton into five parts and feed them into five RNN networks; \cite{shahroudy2016ntu} proposes new RNN structure to model the long-term temporal correlation of the features for each body part; \cite{liu2016spatio} extends RNN to spatio-temporal domains to analyze the hidden sources of action related information within the input data over both domains concurrently; \cite{zhu2016co} proposes fully connected deep LSTM for skeleton based recognition; \cite{zhang2017geometric} selects a set of geometric features to describe joint relations and apply LSTM thereon. Besides RNN and LSTM, CNN is also explored to recognize human action: \cite{li2017skeleton} proposes a two-stream CNN in which one stream's input is the raw coordinates and the other stream's input is motion data obtained by subtracting joint coordinates in each two consecutive frames; \cite{ke2017new} first transforms skeleton sequence into three clips corresponding to three cylindrical coordinates of skeleton sequence, and then applies deep CNN on them; \cite{soo2017interpretable} proposes temporal convolutional neural networks to explicitly learn readily interpretable spatio-temporal representations for 3D human action recognition; \cite{yan2018spatial} constructs a spatial-temporal human skeleton graph first, and then apply CNN networks. Our model is also a CNN based model that utilizes skeletal data. But unlike the existing models, our convolution is performed on edges instead of nodes.
 
\begin{figure}
\centering
\includegraphics[height=6.8cm]{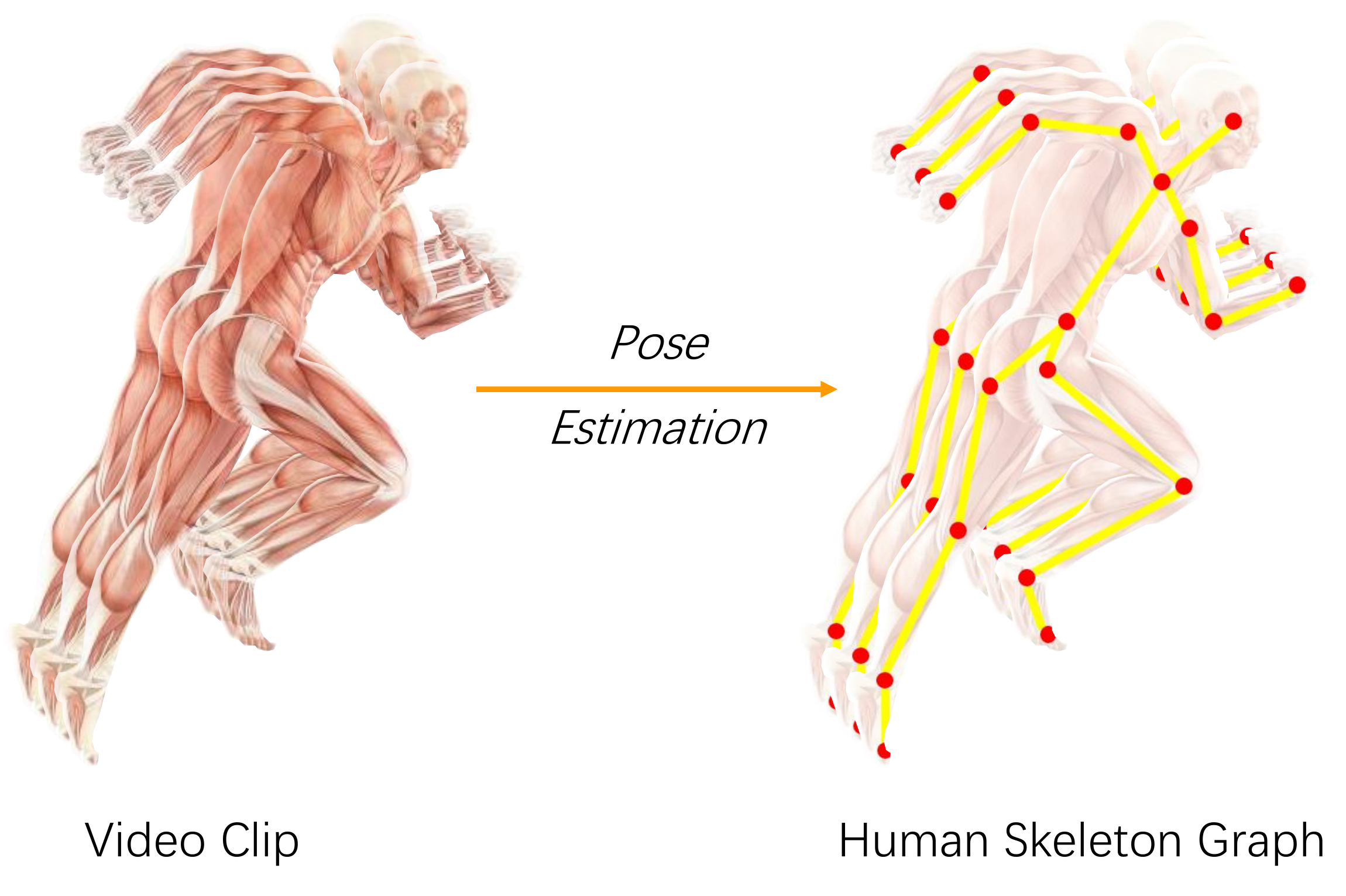}
\caption{The transformation from raw video clip into frames of human skeleton graph. {\it Left} is the raw video clip, and {\it right} is frames of skeleton graph obtained by pose estimation algorithm. {\it Yellow edges} in right figure represent edges (bones), and {\it red dots} represent joints.}
\label{fig:example}
\end{figure}

\section{Graph Edge Convolutional Neural Networks}

In this section, we first review classical convolution operation conducted on images, then we move to our graph edge convolution and graph edge convolutional neural networks as well as its application to skeleton based action recognition. Finally, two hybrid models in which we integrate graph edge convolution and graph node convolution into one single neural network are introduced.

\subsection{Classical Image Convolution}

To make our graph edge convolution more understandable and exhibit its connection with image convolution, we first present a reformulation of classical convolutions on image pixels. Given a gray-scale image, we center the convolution kernel at pixel $x_{ij}$ where $i$ and $j$ denote the row index and column index, respectively. The convolution output can be derived as follows:
\begin{align}
  &x_{ij}^{out} = \sum_{x_{mn} \in {\bf N({\it x_{ij}})} } x_{mn}w(l(x_{mn}))&
\end{align}
where $x_{ij}^{out}$ is the output of this convolution operation, {\bf N}($x_{ij}$) represents the set of neighboring pixels of $x_{ij}$, and $x_{mn}$ denotes the pixels in {\bf N}($x_{ij}$). In practice for a gray-scale image, {\bf N}($x_{ij}$) used to be composed of eight pixels next to $x_{ij}$ and $x_{ij}$ itself, which is exactly the receptive field of this convolution. $l$ is a labeling function that assigns a different order onto each individual element in {\bf N}($x_{ij}$). For example, given a 3$\times$3 convolutional kernel on a gray-scale image, $l(\cdot)$ will assign $\{1, 2, \cdots, 9\}$ respectively to nine pixels in {\bf N}($\cdot$) from left to right and top to bottom.  Weight function $w$ will give each pixel a weight $w(l(\cdot))$ according to the order of this pixel. The value of each pixel is then multiplied with its corresponding weight, and the summarization of these weighted values of pixels is regarded as the convolution output at the center pixel. Similar convolution operation can be applied for RGB color images. Since there are three different channels in a RGB image, the value of each pixel is a three-dimensional feature vector and its corresponding weight becomes a three-dimensional weight vector as well. The multiplication between pixels and weights is extended to inner product.

Convolutional neural networks (CNNs) have received impressive performance in various scenarios, e.g. image classification and object detection \cite{ge2016robust,krizhevsky2012imagenet,redmon2016you,ren2015faster,simonyan2014very,zhang20173,xiong2017learning}. However, it is not straightforward to apply pixel convolution to graph data. This is because graph data does not have the gridded array structure as image, video, and signal data. Each pixel in gridded images would have the same number of neighbors and the same relationships to a neighbor in a given direction. Non-gridded graphs do not have these limitations. A non-gridded graph can vary in the number of neighbors from vertex (edge) to vertex (edge), and there is not necessarily a geometrical interpretation for any given connection between two vertexes (edges).

\subsection{Graph Edge Convolution}

We represent a graph by {\bf G} = ({\bf V}, {\bf E}), in which ${\bf V} = {\it \{v_i | i = 1, ... ,N \} }$ is the node set with $N$ elements, and {\bf E} = \{{\it $e_{ij} | v_i, v_j$ are connected }\} is the edge set containing all edges of {\bf G}. And several concepts have to be introduced to facilitate the explanation of graph convolution.
1) Path between two edges: A path is a set of distinct nodes and edges connecting two edges in a graph (Fig. 3a). More than one path may exist between two edges.
 2) Length of path between two edges:  Given a specific pair of edges, more than one path may exist between the two edges, and these different paths contain different number of nodes and edges. For a certain path, the number of the nodes contained in it is defined as the length of the path. Paths with different lengths between two specific edges are illustrated in Fig. 3b.
 3) Shortest path between two edges: Given a certain pair of edges, among all paths connecting the two edges, the path with the smallest length is defined as the shortest path between the two edges. An illustration of shortest path is given in Fig. 3b.
4) Distance between two edges: Given a certain pair of edges, the length of the shortest path is defined as the distance between these two edges, denoted as {\bf D($\cdot,\cdot$)}. An illustration of distance between a specific pair of edges is given in Fig. 3b.

\begin{figure}
\centering
\includegraphics[height=5.3cm]{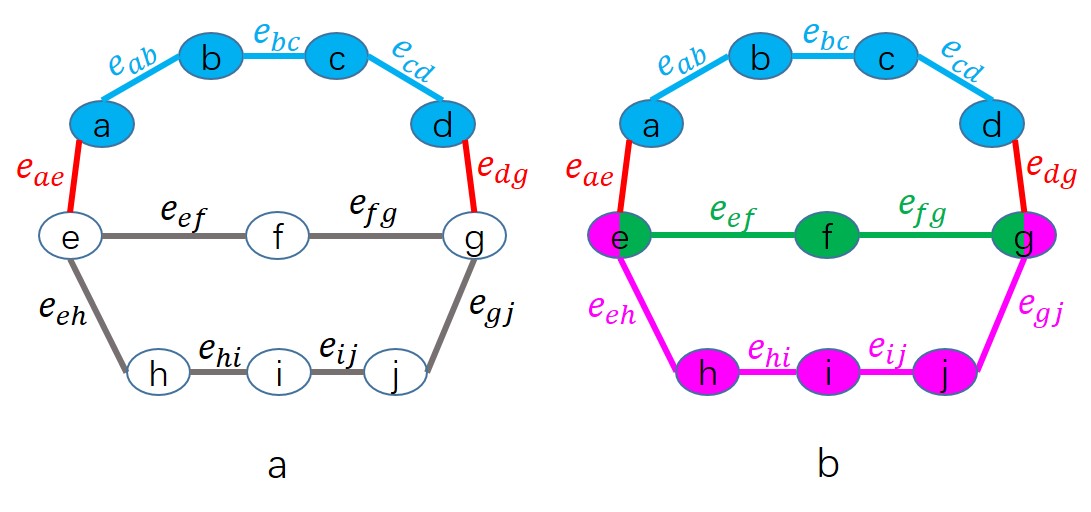}
\caption{The blue parts (node $a$, $b$, $c$, $d$ and edge $e_{ab}$, $e_{bc}$, $e_{cd}$) in left figure illustrates a path between $e_{ae}$ and $e_{dg}$, and according to the definition, the length of this path is 4. In the right figure, the parts in blue, green, and mauve depict three different paths between $e_{ae}$ and $e_{dg}$. As node $e$ and $g$ are shared by two paths, thus they have two colors. The lengths of blue path, green path, and mauve path are 4, 3, and 5 respectively. The length of green path is the smallest, thus it's the shortest path between $e_{ae}$ and $e_{dg}$, and the length of the green path, which is 3, is also the distance between $e_{ae}$ and $e_{dg}$.}
\label{fig:example}
\end{figure}

To conduct convolution operation, we need a receptive field, which is also called a neighborhood in the following. Here we define the neighborhood of a certain edge $e_{pq}$ as 
{\bf  N({\it $e_{pq}$})} = \{{\it $e_{kl}$ $|$ $e_{kl}$ $\in$ {\bf E} and {\bf D}$(e_{pq}$, $e_{kl})$$\leqslant$R}\}, where {\bf D}$(e_{pq}$, $e_{kl})$ is the distance between root edge $e_{pq}$ and edge $e_{kl}$, $R$ is a integer denoting the maximal distance between root edge and neighboring edges.   Thus we can also express the neighborhood of a root edge as: edges within a certain distance $R$ from root edge. In our convolution, we set $R$ as 1 (Fig. 4a), and the resulting neighborhood only contains the edges that are directly connected with root edge by one node, i.e. in our convolution, the neighborhood of a certain edge $e_{pq}$ is {\bf  N({\it $e_{pq}$})} = \{{{\it $e_{kl}$ $|$ $e_{kl}$ $\in$ {\bf E} and {\bf D}$(e_{pq}$, $e_{kl})$$\leqslant$} 1}\}.

\begin{figure}
\centering
\includegraphics[height=5.3cm]{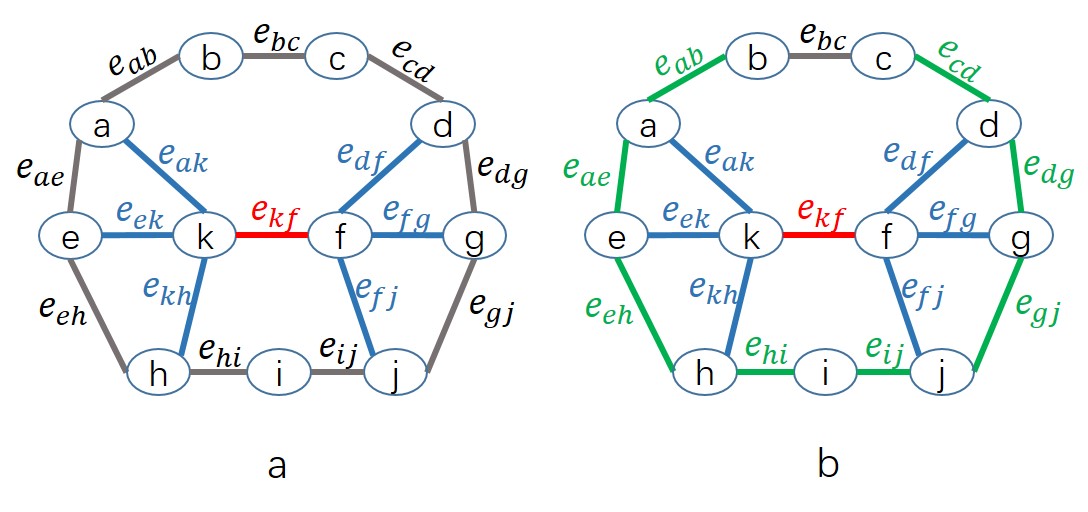}
\caption{In these two figures, the root edge is $e_{kf}$, and the neighboring edges are colored blue or green. The blue edges in left figure are neighboring edges when $R=1$ The blue and green edges in right figure are neighboring edges when $R=2$, while blue and green ones are edges with distances of 1 and 2 from root edge $e_{kf}$ respectively.}
\label{fig:example}
\end{figure}

The convolution at a certain edge $e_{pq}$ is a weighted summation on all the neighboring edges. Formally, the output of convolution operation at edge $e_{pq}$ can be written as (Fig. 1):
\begin{align} 
&e_{pq}^{out} = \sum_{e_{kl} \in {\bf N({\it e_{pq}})}} e_{kl} \cdot w(l(e_{kl}))&
\end{align}

The  $e_{pq}^{out}$ on the left of eq.2 is the output of convolution conducted at edge $e_{pq}$, and on the right of the equation is a weighted summation, in which $e_{kl}$ represents all neighboring edges being summed and $w(l(e_{kl}))$ denotes the weight assigned to edge $e_{kl}$. 

As the neighboring edges in a graph do not have an implicit spatial order that neighboring pixels in an image have, to assign weights onto neighboring edges, we have to firstly assign an order on them, and then match each edge with a weight value according to its order. The labeling function $l$ is just the function that assigns order on the neighboring edges. For each edge $e_{kl}$ in the neighborhood, the labeling function $l$ will assign a labeling value $l(e_{kl})$ on it, denoting the order of this edge, and the weight assigned to $e_{kl}$ is depended on the labeling value $l(e_{kl})$. Unlike in the images in which each pixel has a fixed number of neighboring pixels, edges in graphs do not have a fixed number of neighboring edges. Thus we cannot prepare a fixed number of weights to assign on edges as the number of neighboring edges may vary. So, instead of giving each neighboring edge a unique labeling value, our labeling function $l$ will map neighboring edges into a fixed number of subsets, and edges in same subset will have same labeling value. i.e. $l$: {\bf N}($e_{pq}$)$\to$\{1,...,$K$\}, every single edge in the neighborhood will be labeled an integer ranging from 1 to $K$, and this integer is the edge's order that determines which weight value is going to be assigned to this edge. Thus we see, even if the number of neighboring edges is not fixed, we can always assign them with $K$ weight values because the edges will always be divided into $K$ subsets. $w$ in eq.2 is weight function that assigns weights to edges according to their labeling values, and the edges with same labeling values will be given a same weight. After being multiplied with weights, the summation is conducted on all of the neighboring edges and the convolution operation is completed by now.

Similar to image convolution mentioned above, by replacing edge feature and weight with edge feature vector and weight vector, we can easily extend our model to graphs with multiple channels.

To balance the contribution of edges with different weights, we add a normalizing term in our eq.2
\begin{align}
&e_{pq} = \sum_{e_{kl} \in {\bf N({\it e_{pq}})}} \frac{1}{Z_{pq}(e_{kl})} e_{kl} \cdot w(l(e_{kl})) &
\end{align}

{${Z_{pq}(e_{kl})}$} is the number of neighboring edges with the same labeling value as $e_{kl}$ in edge $e_{pq}$ 's neighborhood, i.e. 
\begin{align}
&Z_{pq}(e_{kl}) = |  \{e_{mn}|e_{mn} \in{\bf N}(e_{pq}) \  and \  l(e_{mn})=l(e_{kl})\}  &
\end{align}

Thus we can see, when the neighboring edges are not divided evenly into $K$ subsets, {\large $\frac{1}{Z_{pq}(e_{kl})}$} is the term  to balance the contribution of neighboring edges with different labeling values. By now, the convolution operation of edges has been elucidated, and we can build our Graph Edge Convolutional Neural Networks using it.

\subsection{Graph Edge Convolutional Neural Networks}
For each video clip containing human actions, the extracted skeletal data are stored as a sequence of frames. In each frame, human bodies are represented as skeleton graphs, in which nodes denote human joints and edges denote human bones.
Our edge convolution can be directly conducted on a single frame of human skeleton graph. But in human action recognition, a sequence of human skeleton graphs has to be dealt with, and skeleton graphs in different frames have to be processed concurrently to capture the dynamics of human body. Thus we have to redefine the neighborhood of an edge to incorporate edges in different frames to get our convolution conducted in both spatial and temporal domains. 

For a certain edge $e_{pq}$ in frame $\tau1$, we denote it as $e_{pq}^{\tau 1}$, and the neighboring edges of $e_{pq}^{\tau 1}$ is defined from two aspects: 1) in frame $\tau 1$, edges connected to $e_{pq}^{\tau 1}$ via only one node are its {\it spatial neighbors}, which is just the definition of neighborhood mentioned above; 2) in another frame $\tau2$, if $\tau2$ is within a certain distance from $\tau1$ i.e. $| \tau 2 $ - $\tau 1 |<K_t$, the {\it spatial neighbors} of $e_{pq}^{\tau 2}$ is also regarded as neighbors of $e_{pq}^{\tau 1}$ and are called {\it temporal neighbors} of $e_{pq}^{\tau 1}$, i.e. \\
 {\bf N}($e_{pq}^{\tau 1}$)=\{ $e_{kl}^{\tau 2}$ $|$ $e_{kl}^{\tau 2}$ $\in$ {\bf E} and $| \tau 2 $ - $\tau 1 |$ $\leqslant$ $K_t$ and {\bf D}($e_{kl}^{\tau2}$, $e_{pq}^{\tau2}$) $\leqslant$ 1\}\\

\iffalse
\begin{align}
& {\bf N}(e_{pq}^{\tau 1})=\{e_{kl}^{\tau 2}|e_{kl}^{\tau 2} \in {\bf E} and | \tau 2  - \tau 1 | \le K_t \quad and \quad {\bf D}(e_{kl}^{\tau2}, e_{pq}^{\tau2}) \leqslant 1             \} &
\end{align}
\fi

%用数学符号写一下第五面第二段的计算
%混合模型的也是，两个stream是两个函数，能想到的用数学语言的都用一下，混合模型名字起一个low level和high level模型
Here $\tau 1$ and $\tau 2$ denote two different frames. 
$K_{t}$ is an integer called temporal kernel size, which is to restrict that temporal neighbors of $e_{pq}^{\tau 1}$ must not be more than $K_{t}$ frames away. And the constraint {\bf D}($e_{kl}^{\tau2}$, $e_{pq}^{\tau2}$) $\leqslant$ 1 is to define spatial neighbors.

The labeling function defined in section Graph Edge Convolution is for spatial neighbors in a single frame. Considering the temporal neighbors, labeling function is rewritten as:
\begin{align}
&l'(e_{kl}^{\tau 2}) = l(e_{kl}^{\tau 2})+(\tau 2  - \tau 1 + K_t) \times K&
\end{align}
Here $l'$ is the rewritten labeling function, $l$ is labeling function in a single frame for spatial neighbors, $K_t$ is temporal kernel size, and $K$ is the number of subsets divided by labeling function $l$ mentioned in section Graph Edge Convolution. Moreover, adding $K_t$ to $\tau1-\tau2$ is to ensure (${\tau2-\tau1+K_t}$) is nonnegative, and multiplying $K$ at the end is to ensure the labeling values for temporal neighbors are not same as spatial neighbors.

As our model is to convolve edges instead of nodes and the original skeletal data are coordinates of human joints (nodes), we have to firstly transform the skeletal data from node form into edge form. The data of each bone (edge) is calculated from coordinates of the two joints on the bone's end. For each bone, we first get the coordinates of the bone's center by averaging the coordinates of the joints on two ends, and then subtract one joint's coordinates from the other to obtain a vector whose length and orientation represent the length and orientation of the bone between the two joints. Denoting the bone (edge) of which the information is being calculated as $e_{pq}$, the two joints (nodes) on the bone's two ends are $n_p$ and $n_q$, we can formally write the calculation of the center coordinates of $e_{pq}$ as:
\begin{align}
&x_c(e_{pq}) = \frac{1}{2} \times (x(n_p) + x(n_q))&\\
&y_c(e_{pq}) = \frac{1}{2} \times (y(n_p) + y(n_q))&\\
&z_c(e_{pq}) = \frac{1}{2} \times (z(n_p) + z(n_q))&
\end{align}
in which $x_c(e_{pq})$, $y_c(e_{pq})$, and $z_c(e_{pq})$ denote three coordinates of the center of bone $e_{pq}$, $x(n_p)$, $y(n_p)$ and $z(n_p)$ are coordinates of node $n_p$, which also applies to node $n_q$. And the orientation vector whose length and orientation denote the length and orientation of bone $e_{pq}$ can be formulated as:
\begin{align}
&Ori(e_{pq}) = (x(n_p) - x(n_q), y(n_p) - y(n_q), z(n_p) - z(n_q))&
\end{align}

So we see that each bone is represented by a set of coordinates denoting the center of the bone and a vector denoting the bone's length and orientation.

Motions of different human body parts can be roughly categorized as concentric and eccentric, thus we choose our labeling function $l$ for spatial neighbors as {\it spatial configuration labeling}. Specifically, for convolution at a certain edge $e_{pq}$, this labeling function divides neighboring edges into three subsets: 1) edges that are closer to gravity center than $e_{pq}$; 2) edges with an equal distance to gravity center as $e_{pq}$; 3) edges that are farther to gravity center than $e_{pq}$. i.e.:
\begin{equation} l(e_{kl})=
\begin{cases}
0 \quad if \quad {\bf d}(e_{kl}, G_c) = {\bf d}(e_{pq}, G_c)& \\
1 \quad if \quad {\bf d}(e_{kl}, G_c) < {\bf d}(e_{pq}, G_c) & \\
2 \quad if \quad {\bf d}(e_{kl}, G_c) > {\bf d}(e_{pq}, G_c)
\end{cases}
\end{equation}

\begin{figure*}
\centering
\includegraphics[height=4.7cm]{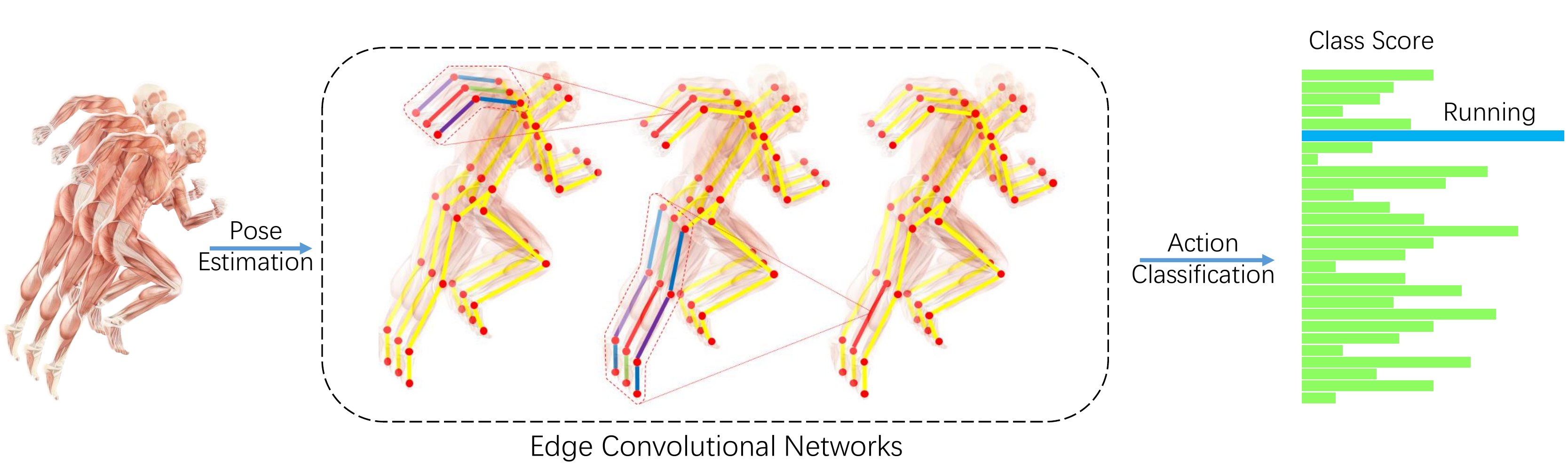}
\caption{Illustration of the edge convolution system. The {\it middle figure} illustrates the process of edge convolution, the edges to be convolved are colored, and different colors indicate different labeling values.}
\label{fig:example}
\end{figure*}

In eq.11, $G_c$ is the gravity center of human body calculated by averaging coordinates of all body parts, and {\bf d}($e_{kl}$, $G_c$) denotes the distance from edge $e_{kl}$ to $G_c$. Having the labeling function $l$ for spatial neighbors, the labeling function for temporal neighbors can be formulated according to eq.5.

\subsection{Combining Graph Edge and Node Convolutions}
 
A convolutional network is to extract a set of high level features from the original skeleton sequence, but the node convolutional network and the edge convolutional network extract features from different perspectives, thus the two sets of features (output of the convolutional layer) represent a same video sequence from different viewpoints. Both edge convolution and node convolution have their unique advantages that cannot be replaced by each other. Just like we mentioned before, the edge convolutional network leverages the dynamics of bones, while the node convolutional network leverages dynamics of joints. 

Thus we wonder what if we combine them together to get hybrid models that can utilize advantages of both edge and node convolution? As dynamics of joints and bones are complementary to each other, we guess that designing a model to utilize the two sets of features together will enable us to leverage the human skeleton dynamics from both joint-perspective and bone-perspective and further improve the performance on human action recognition task. 
So we design two different hybrid models that combine edge and node convolution models by features of different levels.

1) Sequence level hybrid model: 
A straight forward realization of this idea is our sequence level hybrid model. This is a two-stream model. In one stream we perform layers of edge convolution onto the skeleton sequence and in the other stream we perform layers of node convolution. Given the input sequence $s_{in}$, the output of the convolutional layer in the edge convolution stream can be formulated as:
\begin{align}
& h_{seq}^e = f_{conv}^e (s_{in})&
\end{align}

in which $f_{conv}^e$ represents layers of edge convolution, and $h_{seq}^e$ is the output from the last convolutional layer of the edge convolution stream. And the output of the convolutional layer in the node convolution stream can be formulated as:
\begin{align}
& h_{seq}^n = f_{conv}^n (s_{in})&
\end{align}

in which $f_{conv}^n$ represents layers of node convolution, and $h_{seq}^n$ is the output from the last convolutional layer of the node convolution stream.

In the sole edge convolution or node convolution networks, we have only one set of features $h_{seq}^e$ or $h_{seq}^n$ and we apply a global pooling on it to get a representation for the whole sequence, and then input it into a fully connected layer 
%（改成fully connected layer）
 to output final class scores denoting the probabilities for the sequence to be classified into each class. But now we have two sets of features and two different representations for a same sequence. What we do is to concatenate these two representations into a single tensor to get the input for the last fully connected layer, i.e.
\begin{align}
& I_{seq}^{fc} = f_{pool}(h_{seq}^e) \oplus f_{pool}(h_{seq}^n)&
\end{align}
in which $I_{seq}^{fc}$ denotes the input to the fully connected layer in sequence level hybrid model, $f_{pool}$ is the global pooling layer, and $\oplus$ is a concatenation operator.
And the final output of the whole model is:
\begin{align}
& O_{seq} = f_{fc}(I_{seq}^{fc})&
\end{align}
in which $f_{fc}$ denotes the fully connected layer, and $O_{seq}$ is the final output of the sequence level hybrid model.

By concatenating the outputs from edge and node convolution streams, features extracted from both edge and node convolutional networks contribute to final classification result, i.e. dynamics of both joints and bones are leveraged in our classification.

In this model, the input to the shared layer are features of the whole skeleton sequence, thus we call the first hybrid model as sequence level hybrid model.

2) Body-part level hybrid model: In sequence level hybrid model, two sets of features complement each other in a shared fully connected layer after being separately extracted by node and edge convolutional network. But we also want the two convolutional networks to supplement each other during the feature extracting period rather than after it, i.e. we want to merge the two networks in a convolutional layer rather than a fully connected layer. Thus we design another hybrid model, and the most distinctive part of this model is the shared convolutional layer.

 In a shared convolutional layer, our convolution operation is neither edge convolution nor node convolution, but a combination of both. As mentioned before, we conduct convolution on a human skeleton graph, when we perform edge convolution, the data are information of bones and attached to edges, and when we perform node convolution, data are information of joints and attached to nodes. But in a shared convolutional layer, both nodes and edges are attached with information, thus when conducting convolution, information of edges can be passed to nodes, and vice versa. By this kind of convolution, dynamics of bones and joints can supplement each other during the period of feature extraction in the shared convolutional layer. An illustration of convolution operation in the shared convolutional layer can be found in Fig. 6.

Similar to sequence level hybrid model, the first part of body-part level hybrid model are two separated streams including an edge convolutional network and a node convolutional network, and the output of node and edge convolutional networks are two sets of features denoted as $h_{body}^e$ and $h_{body}^n$, which can be formulated as follows:
\begin{align}
& h_{body}^e = f_{conv}^e (s_{in})&
\end{align}
\begin{align}
& h_{body}^n = f_{conv}^n (s_{in})&
\end{align}
in which the $f_{conv}^n$, $f_{conv}^e$, and $s_{in}$ have exactly the same meanings as we mentioned in sequence level hybrid model.
 Compared to the sequence level hybrid model, the subscripts are changed from $seq$ to $body$ to indicate that they are parts of body-part level hybrid model. More specifically, the feature set extracted by edge convolutional network contains feature vectors for each bone in each frame, and the feature set extracted by node convolutional network contains feature vectors for each joint in each frame. After getting these two feature sets, we construct a new human skeleton graph, then we fill in each edge with the corresponding feature vector extracted by the edge convolutional network and fill in each node with the corresponding feature vector extracted by the node convolutional network. In this new human skeleton graph, both edges and nodes are filled with feature vectors, which are the high level features of bones and joints containing the dynamic information extracted by the two separated convolutional networks. Then the shared convolutional layer is applied thereon and the input can be formulated as follows:
\begin{align}
& I_{body}^{s-conv} = h_{body}^e \oplus h_{body}^n&
\end{align}
in which $I_{body}^{shared-conv}$ denotes the input to shared convolutional layer in the body-part level hybrid model. In the shared convolutional layer, the high level features of bones and joints communicate with each other and get furnished by each other. The output of the shared convolutional layer is also feature vectors of bones and joints like the output we get from the two separate streams. The difference is that in the two separate streams, the features of joints and bones are separately extracted, but in our shared convolutional layer, each time we perform convolution operation on the human body graph, the information of joints get chances to flow to the neighboring bones, thus the features of bones are informed and get trimmed by dynamics of joints, and vice versa, making the final features more representative. We can formally write it as:
\begin{align}
&h_{body}^{s-conv} = f_{s-conv}(I_{body}^{s-conv}) &
\end{align}
in which $h_{body}^{s-conv}$ represents the output from a shared convolutional layer in the body-part level hybrid model, and $f_{s-conv}$ denotes the shared convolutional layer.
After two layers of shared convolution, we apply a global pooling layer and a fully connected layer to get final classification result just like we do in sequence level hybrid model:
\begin{align}
&h_{body}^{pool} = f_{pool}(I_{body}^{s-conv}) &
\end{align}
\begin{align}
&I_{body}^{fc} = h_{body}^{pool} &
\end{align}

$h_{body}^{pool}$ is the output from global pooling layer and $I_{body}^{fc}$ is the input to the final fully connected layer. The final output from the fully connected layer can be formulated as:
\begin{align}
&O_{body} = f_{fc}(h_{body}^{pool}) &
\end{align}
in which $O_{body}$ is the output of the whole body-part hybrid model, and $f_{fc}$ denotes the fully connected layer.

In our hybrid models, the high-level features extracted from both edge convolution and node convolution can corroborate each other. As we mentioned above, in a lot of situations, the movements of the joints are too subtle to capture, but the movements of bones are always obvious, in these cases, the hybrid models can utilize the power of edge convolution to classify the human actions. However, among some classes of human action, the movements of bones may be too similar to be distinguished from each other, in these cases, the dynamics on joints may help the hybrid model to do the classification. That is to say, dynamics of both joints and bones are leveraged and fused together to recognize human action, thus the skeletal data is fully utilized in our hybrid models.

In our second hybrid model, the features input into the shared layers are features of each specific human body joint and bone, which are body-part level features. Thus we name this model as body-part level hybrid model.

\begin{figure}
\centering
\includegraphics[height=6cm]{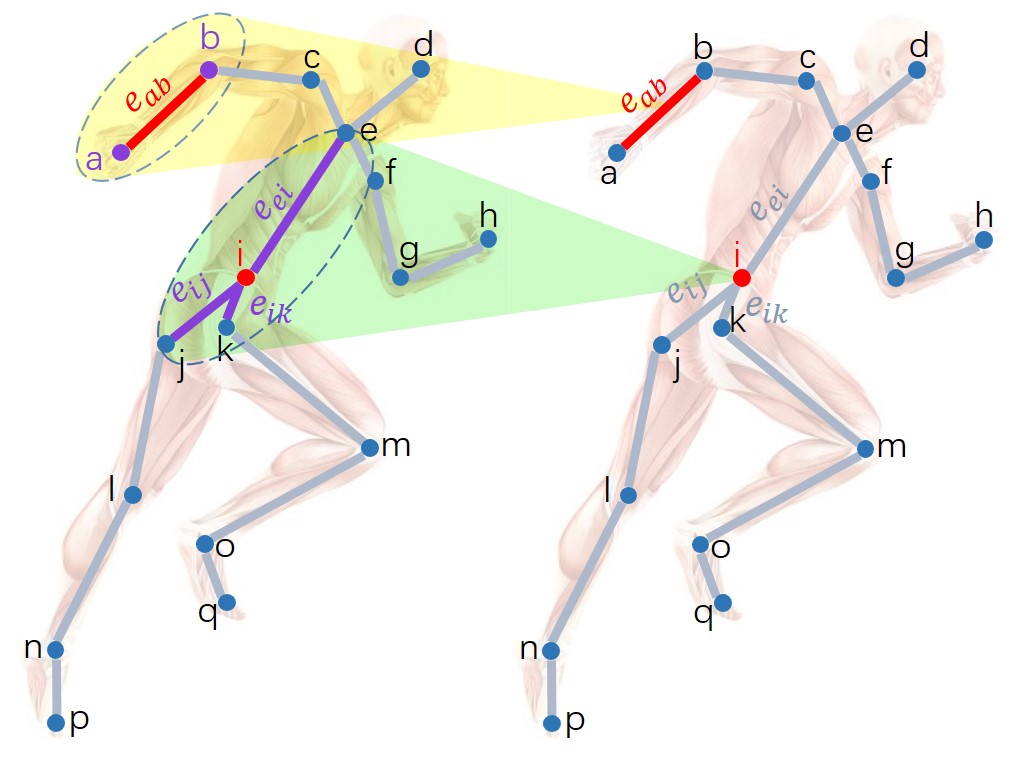}
\caption{Illustration of the convolution in the shared convolutional layer. It shows a convolution at edge $e_{ab}$ and a convolution at node $i$. For the convolution at $e_{ab}$ (with light yellow background), node $a$, $b$ and edge $e_{ab}$ are involved, and a weight summation of them are assigned to the kernel center $e_{ab}$ in next layer. For the convolution at node $i$ (with light green background), edge $e_{ei}$, $e_{ij}$, $e_{ik}$, and node $i$ are involved, a weight summation of them are assigned to the kernel center node $i$ in next layer. In both convolutions, the center of the convolution kernel is colored red, like $e_{ab}$ and node $i$, and the other parts involved are colored purple.} 
\label{fig:example}
\end{figure}

\section{Experiments}
Firstly, we conduct ablation study on Kinetics dataset to explore the influence of temporal kernel size on action recognition and choose a proper temporal kernel size, then we evaluate our model's performance on two datasets: {\bf Kinetics human action dataset} (Kinetics) \cite{kay2017kinetics}, which is by now the largest unconstrained action recognition dataset, and {\bf NTU-RGB+D} dataset \cite{shahroudy2016ntu} the largest in-house captured human action recognition dataset. Then we compared our performance with previous state-of-the-art algorithms. On Kinetics, we compared our results with feature encoding approach \cite{fernando2015modeling}; Deep LSTM \cite{shahroudy2016ntu}; Temporal ConvNet \cite{soo2017interpretable}; ST-GCN \cite{yan2018spatial}. On NTU-RGB+D, we made comparision with Lie Group model \cite{vemulapalli2014human}; H-RNN \cite{du2015hierarchical}; Deep LSTM \cite{shahroudy2016ntu}; PA-LSTM \cite{shahroudy2016ntu}; ST-LSTM + TS \cite{liu2016spatio}; Temporal Conv \cite{soo2017interpretable}; C-CNN + MTLN \cite{ke2017new}; and ST-GCN \cite{yan2018spatial}. To further analyze the improvement brought by our edge convolution, we visualized the classification accuracies for each class on NTU-RGB+D dataset, and compared the accuracy between node convolution model and edge convolution model on some representative classes.
\subsection{Datasets and Settings}

\subsubsection{Kinetics}
Kinetics actually is a dataset containing raw videos retrieved from YouTube. It consists of 300,000 video clips, and all the samples fall into 400 classes with at least 400 video clips for each action class. To get the skeletal data, \cite{yan2018spatial} used $OpenPose$ toolbox \cite{cao2017realtime} to obtained 2D coordinates of joints in the video. Each joint located by $OpenPose$ is in the form of $X, Y, C$, and each body is represented by 18 joints. (Fig. 7) $X$ and $Y$ are two coordinates of the joint, while $C$ is the confidence score. For each video clip, two bodies with the highest average joint confidence score is reserved. Thus each video is represent by a tensor of ($3,T,18,2$), where 3 denotes ($X,Y,C$), $T$ denotes number of frames in the video, 18 is the number of joints of each body, and 2 is number of bodies in the video.

\begin{figure}
\centering
\includegraphics[height=6.6cm]{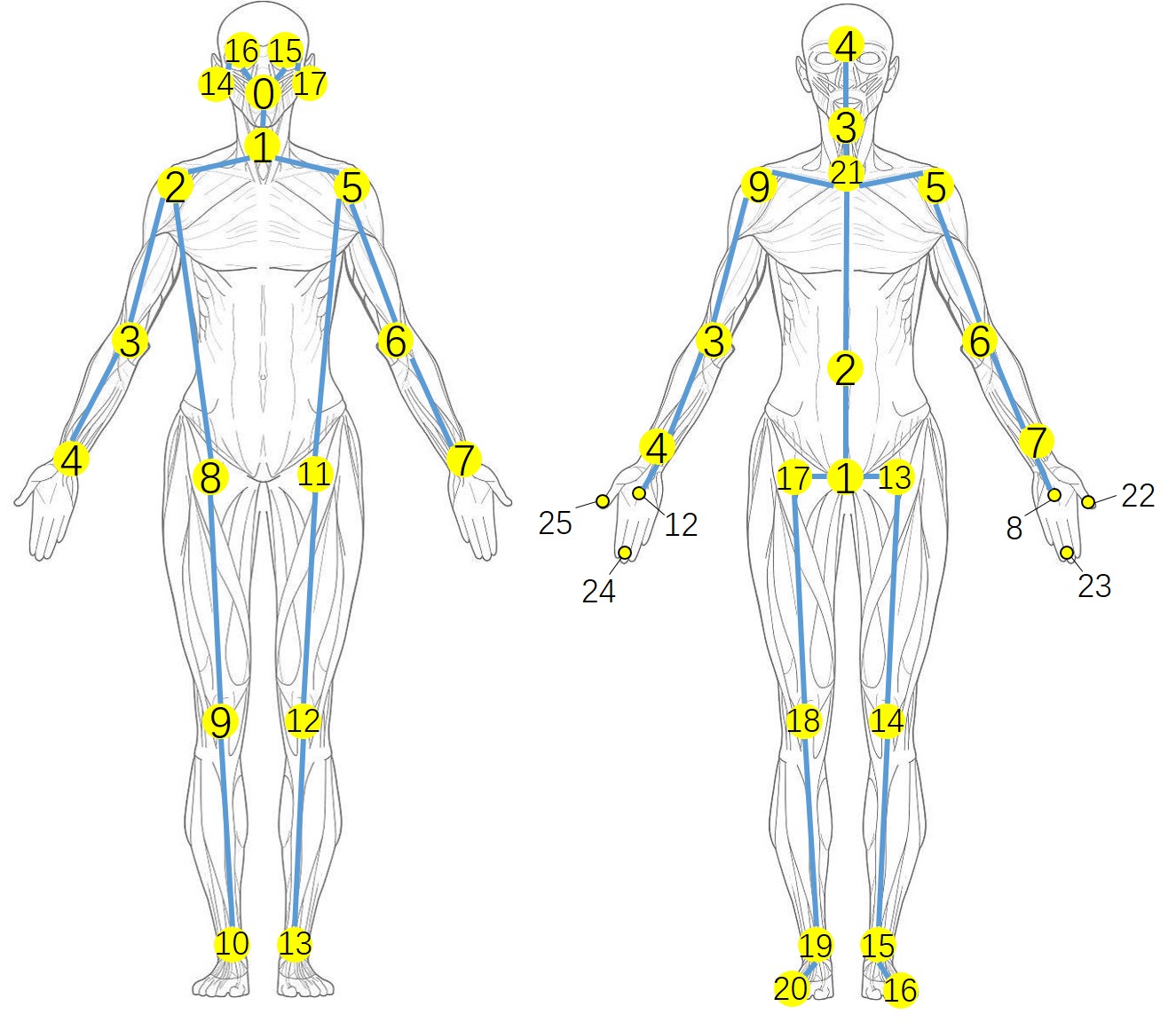}
\caption{Illustration of the human skeleton graphs in two datasets. The figure on the left depicts 17 located human joints and the connections between them in Kinetics dataset, and the figure on the right containing 25 joints is for NTU-RGB+D dataset.}
\label{fig:example}
\end{figure}

As mentioned before, we have to transform the skeletal data from joint form into bone form. In this dataset, the original skeletal data is 2D coordinates and confidence score of joints, and our human skeleton graph is composed of 18 joints and 17 bones. We represent each bone by five values: the first two values are 2D coordinates of the bone center obtained by averaging the coordinates of the joints on the bone's two ends, the third value corresponds to the confidence score of the bone obtained by averaging the confidence scores of the two joints on the bone's two ends, and the last two values are components of a 2-dimension vector (orientation vector) obtained by subtracting the coordinates of the joints on the bone's ends that denotes the length and orientation of the bone.

The whole skeletal dataset retrieved from Kinetics contains 266,440 samples. 246,534 of the dataset are used for training and 19,906 samples are used for testing in our setting. And on Kinetics dataset, we report both top-1 and top-5 accuracies.

\subsubsection{NTU-RGB+D}
By now, this is the largest dataset with 3D joint annotations for human action recognition task. The dataset is captured by 3 Microsoft Kinetic v.2 cameras concurrently by \cite{shahroudy2016ntu}. It contains 56,880 action samples in form of RGB video, depth map sequences, 3D skeletal data, and infrared videos. All samples in this dataset fall into 60 action classes, Among the multiple modalities in this dataset, the skeletal data is the only one we use. In the skeletal data, each human body is represented by 25 joints (Fig. 7), and each joint is represented by {\it X, Y, Z} coordinates.

For this dataset, we construct a graph containing 25 joints and 24 bones. For each bone, we use a six-dimension feature vector to denote the bone's location and orientation. The first three values of the feature vector are 3D coordinates of the bone center calculated by averaging the 3D coordinates of two joints on the bone's ends and the last three values are 3 components of the orientation vector of the bone, which is obtained by subtracting the coordinates of the two joints on the bone's ends.

The author of this dataset recommend to divide the data in two ways:
{\bf 1) cross-subject} (X-Sub)
In this setting, the data are separated into 40,320 samples and 16,560 samples, where the former part come from one group of actors is used for training, and the latter part come from the other group of actors is used for testing.
{\bf 2) cross-view} (X-view) In this setting, the training set is clips captured by one set of cameras containing 37,920 samples, and the testing set is captured by the other set of cameras containing 18,960 samples. 
On NTU-RGB+D dataset, we only report top-1 accuracy.

\subsection{Pipeline}
\subsubsection{Graph Edge Convolutional Neural Network (GECNN)}
The skeletal data (coordinates of joints) input to our model is first normalized by a batch normalization layer. Then center coordinates and orientation of bones is calculated from the normalized data. Then the data is fed into 9 layers of graph edge convolutional neural networks. The first three convolutional layers have 64 channels for output, the next three layers have 128 channels for output, and the last three layers have 256 channels for output. In the fourth and seventh convolutional layer, the stride is set to 2 to conduct pooling. The output tensor of convolutional layers is then fed into a global pooling layer. The global pooling layer is to get a 256-dimension feature vector for each video sequence. After the global pooling layer, the output tensor is fed into a dense layer to get the class scores for each video sequence. (Fig. 8)

\begin{figure}
\centering
\includegraphics[height=6.9cm]{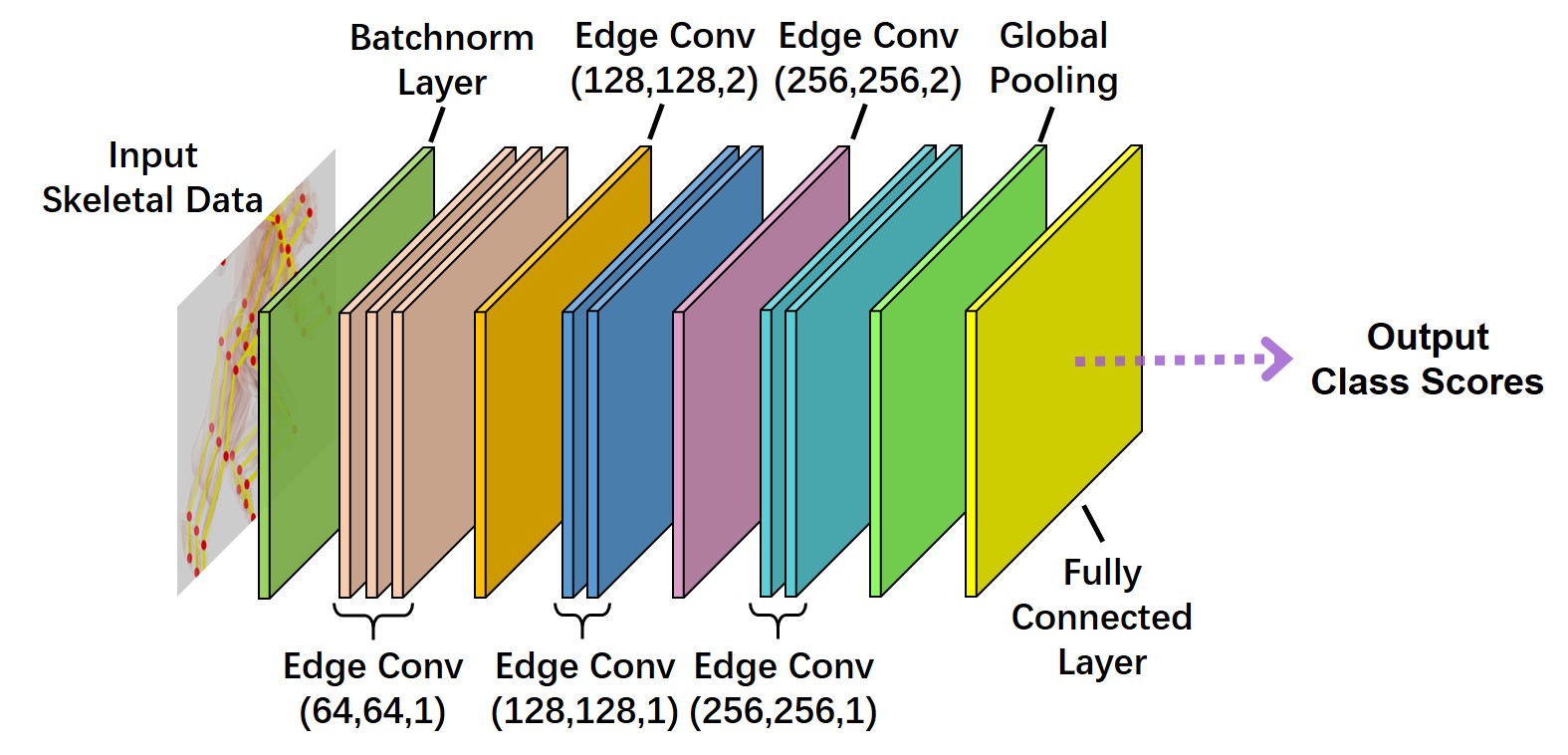}
\caption{Illustration of the network structure of our edge convolution model, the {\it middle} 9 layers are edge convolutional layers. The number of input channels, number of output channels, and stride are denoted as (input, output, stride). e.g. (64, 64, 1) means this layer has 64 input channels, 64 output channels, and the stride is 1.}
\label{fig:example}
\end{figure}

\subsubsection{Sequence level hybrid model}
As mentioned before, the first part of sequence level hybrid model are two separated streams including an edge convolutional stream and a node convolutional stream. The edge convolutional includes a normalization layer, following 9 edge convolutional layers, and a global pooling layer. The node convolution stream has same structure including a normalization layer, 9 node convolutional layers, and a global pooling layer. 

After the global pooling layer in the first part, each sequence is represented as a 256-dimension feature vector. In the second part of sequence level hybrid model, we concatenate the two 256-dimension output tensors from the global pooling layer into a single tensor, and then input it into a fully connected layer to get the final classification result, which is the class scores for each sequence indicating the probabilities of classifying the sequence into each class (Fig. 10).

\begin{figure}
\centering
\includegraphics[height=9cm]{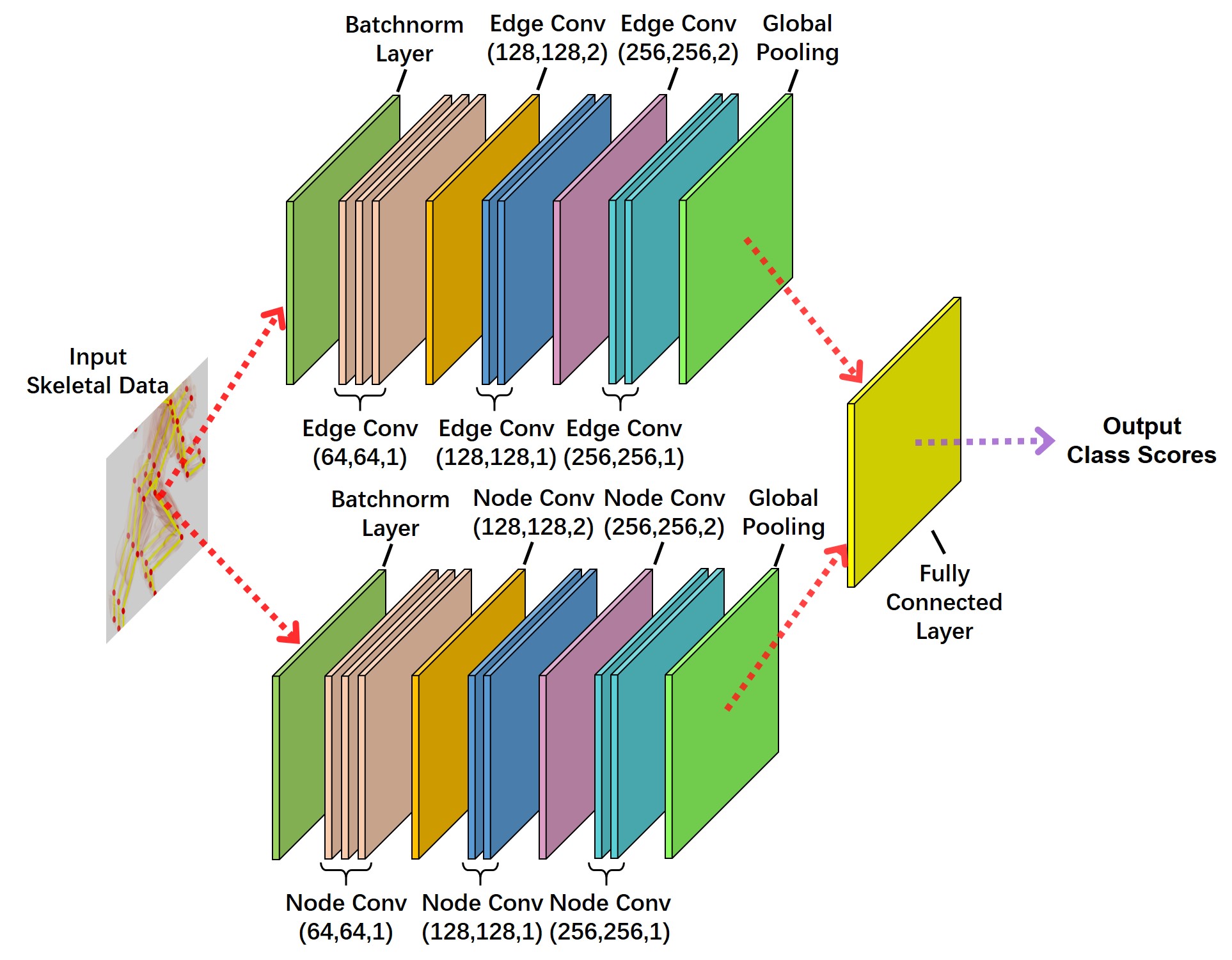}
\caption{Network structure of the sequence level hybrid model. A single skeletal sequence is firstly input into two separated flows, and then the output of the two flows are concatenated and two flows become one.}
\label{fig:example}
\end{figure}

\subsubsection{Body-part level hybrid model}
Hybrid model 2 is also divided into two parts. The first part is same as the first part of body-part level hybrid model, except that the global pooling layer is removed. The second part of body-part level hybrid model includes 4 layers. In the second part, the output tensors of the two former streams are firstly input into two layers of common convolutional layers and then a global pooling is conducted on the output tensors of the common convolutional network, after that, we feed it into a fully connected layer to get the final classification result (Fig. 11).

\begin{figure}
\centering
\includegraphics[height=9cm]{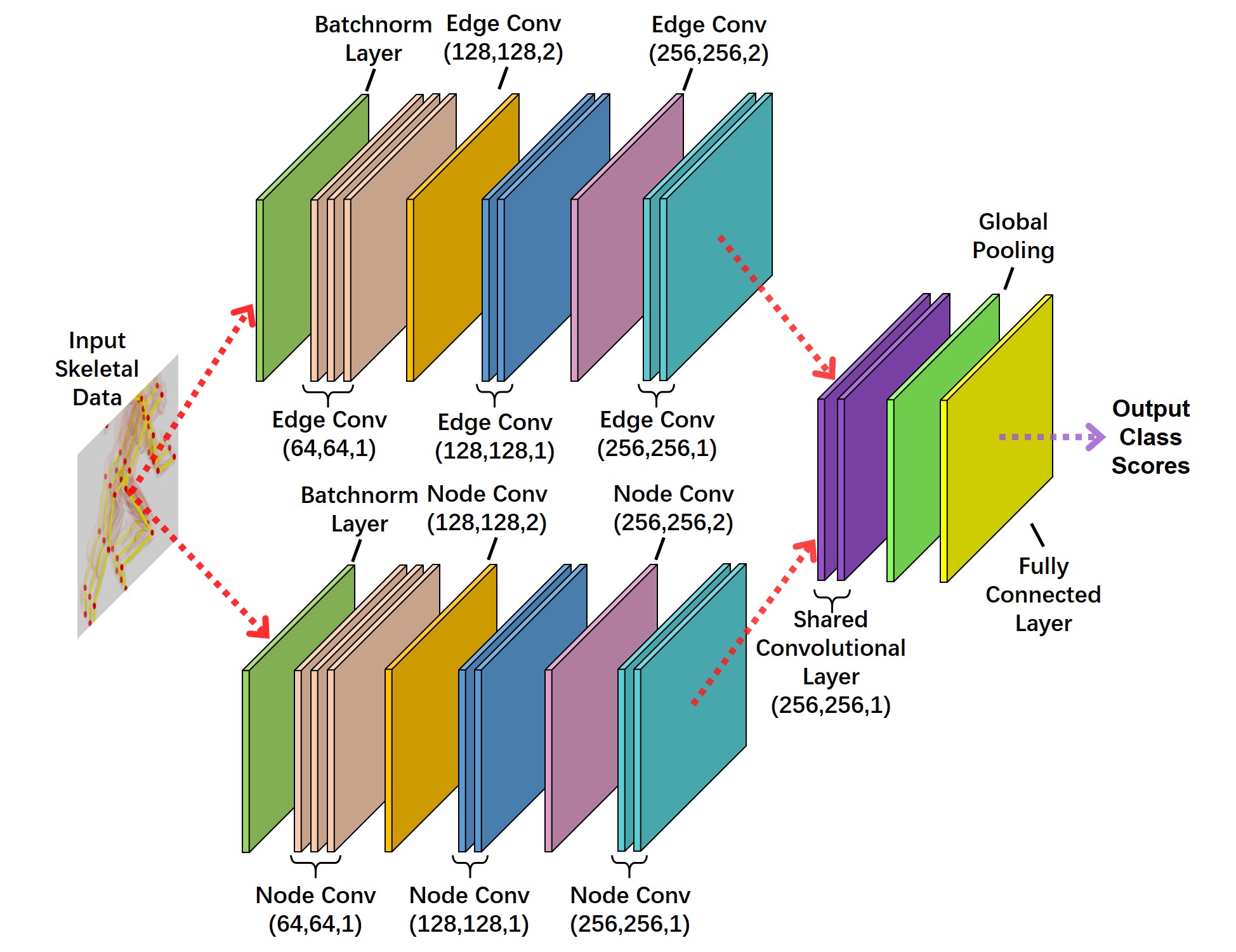}
\caption{Network structure of the body-part level hybrid model.}
\label{fig:example}
\end{figure}

\subsection{Implementation of Edge Convolution}
The implementing of edge convolution actually consists two stages including a temporal-convolution and a spatial-convolution. Each input sequence can be represented as a tensor with shape (C,E,T), in which C denotes the multiple channels of the feature vector, E is the number of edges, and T is the length of the sequence. We first perform traditional 2D convolution with kernel of size 1$\times$$K_t$ on the third dimension to conduct temporal-convolution, and then we multiply the resulting tensor with the adjacency matrix of the human skeleton graph to conduct spatial convolution. The operation to conduct spatial convolution by multiplying the tensor with adjacency matrix is inspired by \cite{sandryhaila2013discrete}, in which the concept {\it graph shift} corresponds to the spatial convolution on graphs.

\subsection{Hyper parameter analysis}
To figure out to what extent incorporating temporal neighbors will improve the performance of action recognition, we choose 6 different temporal size for our edge convolution model and compare the results on Kinetics dataset.

\begin{table}[h]
\begin{center}
\caption{Ablation study on Kinetics datasets. Both top-1 and top-5 accuracies are reported}
\label{table:ablation}
\begin{tabular}{lll}
\hline\noalign{\smallskip}
Temporal kernel size & top-1 & top-5\\
\noalign{\smallskip}
\hline
\noalign{\smallskip}
3  & 29.30\% & 51.71\% \\
5  & 30.58\% & 52.84\% \\
7  & 31.17\% & 53.95\%\\
9  & 31.41\% & 53.90\%\\
11 & 31.49\% & 54.38\%\\
13 & 31.43\%&53.86\%
\\
\hline
\end{tabular}
\end{center}
\end{table}

From Table \ref{table:ablation} we see that when the temporal kernel size rises from 3 to 9, the top-1 accuracy increases significantly, while further enlargement to 11 and 13 doesn't help much. This demonstrates the necessity to incorporate enough number of consecutive frames in a sequence to capture the motion when conducting the convolution. However, incorporating too many frames is not promising and will cost more computation resources. Thus, after comparision, we decide to set temporal kernel size as 9 to get both satisfying accuracy and lower computation resource cost.

\subsection{Experiment Results}

\subsubsection{Kinetics.}
On this dataset, 246,534 samples of the dataset are used for training while 19,906 samples are used for testing, and we report both top-1 and top-5 accuracies.
We compare our graph edge convolutional neural network (GECNN) model, sequence level hybrid model (SLHM), and body-part level hybrid model (BPLHM) with four other characteristic skeleton based models including: feature encoding approach \cite{fernando2015modeling}; Deep LSTM \cite{shahroudy2016ntu}; Temporal ConvNet \cite{soo2017interpretable}; and ST-GCN \cite{yan2018spatial}. In Table \ref{kinetics}, we can see that our model achieves superior performance than these previous approaches.

On this dataset, our edge convolution network achieves superior classification accuracy over previous state-of-the-art models by 0.7\% in top-1 accuracy and 1.1\% in top-5 accuracy.
Our sequence level hybrid model further improves the performance by 0.4\% in top-1 accuracy and 0.3\% in top-5 accuracy. Body-part level hybrid model performs the best, with 2.7\% higher top-1 accuracy and 3.4\% higher top-5 accuracy.

\begin{table}
\begin{center}
\caption{Comparision between different models on Kinetic datasets. Both top-1 and top-5 accuracies are reported}
\label{kinetics}
\begin{tabular}{lll}
\hline\noalign{\smallskip}
Model & top-1 & top-5\\
\noalign{\smallskip}
\hline
\noalign{\smallskip}
Feature Enc. \cite{fernando2015modeling}  & 14.9\% & 25.8\% \\
Deep LSTM \cite{shahroudy2016ntu} & 16.4\% & 35.3\% \\
Temporal ConvNet \cite{soo2017interpretable} \quad \quad \quad \quad & 20.3\% & 40.0\%\\
ST-GCN \cite{yan2018spatial} & 30.7\% & 52.8\%
\\
\hline
{GECNN} & 31.4\% & 53.9\% \\
{SLHM} & {31.8}\% & {54.2}\% \\
{\bf BPLHM} & {\bf 33.4}\% & {\bf 56.2}\% \\       % in epoch 45
\hline
\end{tabular}
\end{center}
\end{table}

\subsubsection{NTU-RGB+D}
On this dataset, we compare our edge convolution model with eight other previous state-of-the-art models including: Lie Group \cite{vemulapalli2014human}; H-RNN \cite{du2015hierarchical}; Deep LSTM \cite{shahroudy2016ntu}; PA-LSTM \cite{shahroudy2016ntu}; ST-LSTM + TS \cite{liu2016spatio}; Temporal Conv \cite{soo2017interpretable}; C-CNN + MTLN \cite{ke2017new}; and ST-GCN \cite{yan2018spatial}. In Table \ref{ntu}, we can see the result and the great improvement compared with former models. On this dataset, only top-1 accuracy is reported.

For this dataset, we divide data in two ways as recommended by author of the dataset. The first setting is cross-subject (X-Sub) in which 40,320 samples coming from one group of actors are used for training and 16,560 samples coming from the other group of actors is used for testing. In this setting, our edge convolution model greatly improves the prediction accuracy by 2.5\% than previous state-of-the-art model. Moreover, our sequence level hybrid model and body-part level hybrid model further improves this by 0.7\% and 1.4\%. Overall, body-part level hybrid model performs the best, whose accuracy is 3.9\% higher than previous state-of-the-art model.

The second setting is cross-view (X-View) in which the training set is 37,920 samples captured by one set of cameras, and the testing set is 18,960 samples captured by the other set of cameras. In this setting, our edge convolution network achieves accuracy that is 1.1\% higher than previous model, and our sequence level hybrid model and body-part level hybrid model further improve it by 0.3\% and 1.7\%. Body-part level hybrid model still performs the best with an accuracy that is 2.8\% higher than the previous state-of-the-art model.
\begin{table}
\begin{center}
\caption{Comparision between different models on NTU-RGB+D dataset. Top-1 accuracy is reported on two seperation of the dataset}
\label{ntu}
\begin{tabular}{lll}
\hline\noalign{\smallskip}
Model & X-Sub & X-View\\
\noalign{\smallskip}
\hline
\noalign{\smallskip}
Lie Group \cite{vemulapalli2014human} & 50.1\% & 52.8\% \\
H-RNN \cite{du2015hierarchical} & 59.1\% & 64.0\% \\
Deep LSTM \cite{shahroudy2016ntu} & 60.7\% & 67.3\%\\
PA-LSTM \cite{shahroudy2016ntu} & 62.9\% & 70.3\%\\
ST-LSTM + TS \cite{liu2016spatio} & 69.2\% & 77.7\%\\
Temporal Conv \cite{soo2017interpretable} & 74.3\% & 83.1\%\\
C-CNN + MTLN \cite{ke2017new} \quad \quad \quad \quad & 79.6\% & 84.8\%\\
ST-GCN \cite{yan2018spatial} & 81.5\% & 88.3\%
\\
\hline
{GECNN} & 84.0\% & 89.4\% \\
{SLHM} & {84.7}\% &{89.7}\% \\
{\bf BPLHM} & {\bf 85.4}\% &{\bf 91.1}\% \\   % in epoch 75 and epoch 60
\hline
\end{tabular}
\end{center}
\end{table}

\begin{figure}[h]
\centering
\includegraphics[height=11cm]{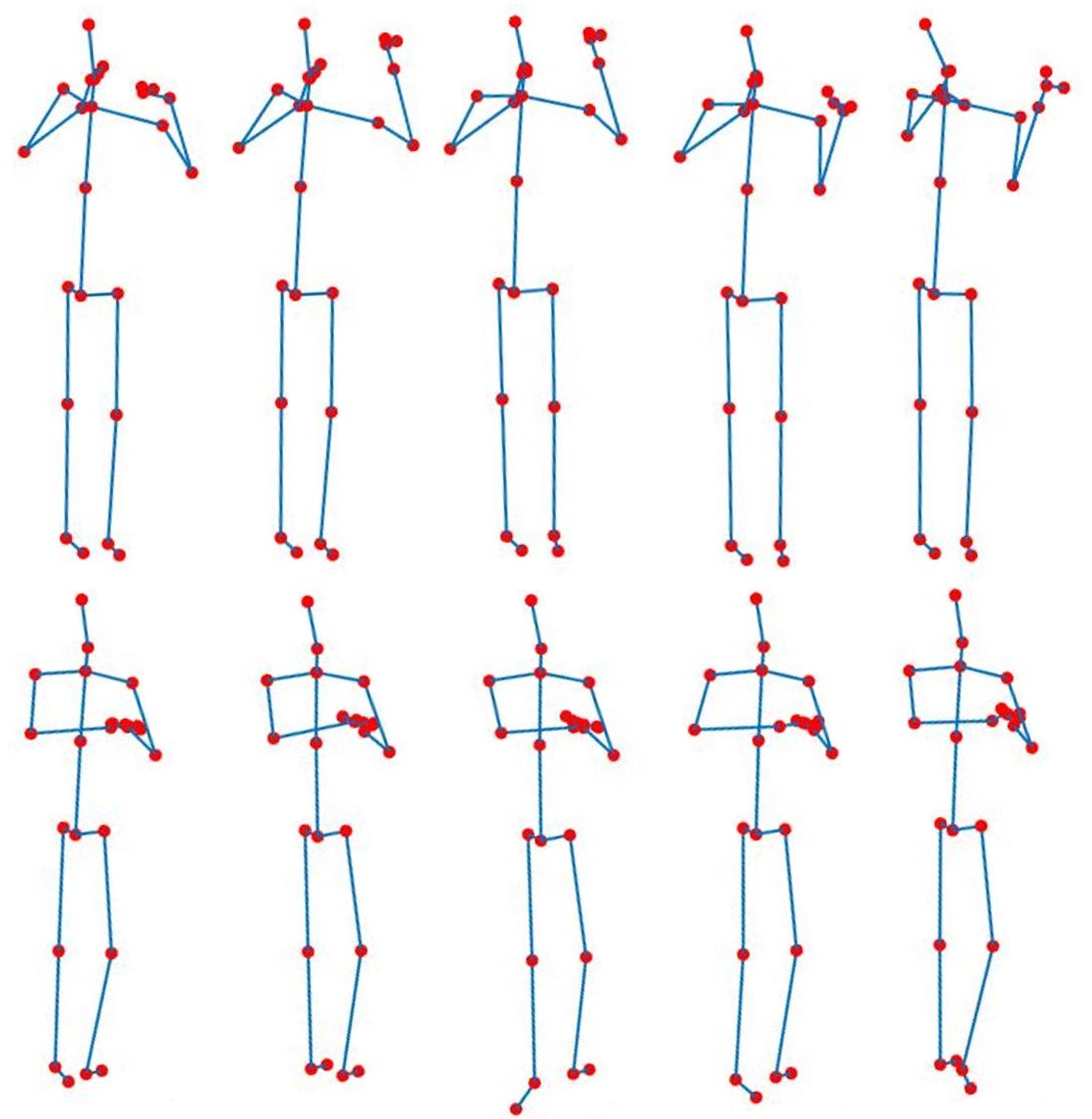}
\caption{The above are 5 frames extracted from an example in the 47th class 'touch neck (neckache)', in which we can see that the joints of hands are seriously overlapped with the joint around neck. The below are 5 frames extracted from the beginning part of an example in the 3rd class 'brushing teeth', and we can see that the joints of two hands overlap with each other greatly.}
\label{fig:example}
\end{figure}

Overall, on all datasets and all different settings, our graph edge convolutional neural network model outperforms the previous state-of-the-art model.
This result preliminarily corroborates our conjecture that in some circumstances, movements of human bones are more obvious than joints, and on average, recognizing human action by analyzing bone dynamics is more efficient than analyzing the joints. Our hybrid models further improve the performance than graph edge convolutional network model, which means that combining edge and node convolution does help improving the performance. The fact that body-part level hybrid model performs better than sequence level hybrid model validates our guess that instead of simply combining the extracted features, merging node and edge convolution models in the feature extracting period enable the two models get more help from each other.
To further validate our analysis, we analyzed the classification accuracies on each class of NTU-RGB+D dataset.

\begin{figure*}
\centering
\includegraphics[height=6.5cm]{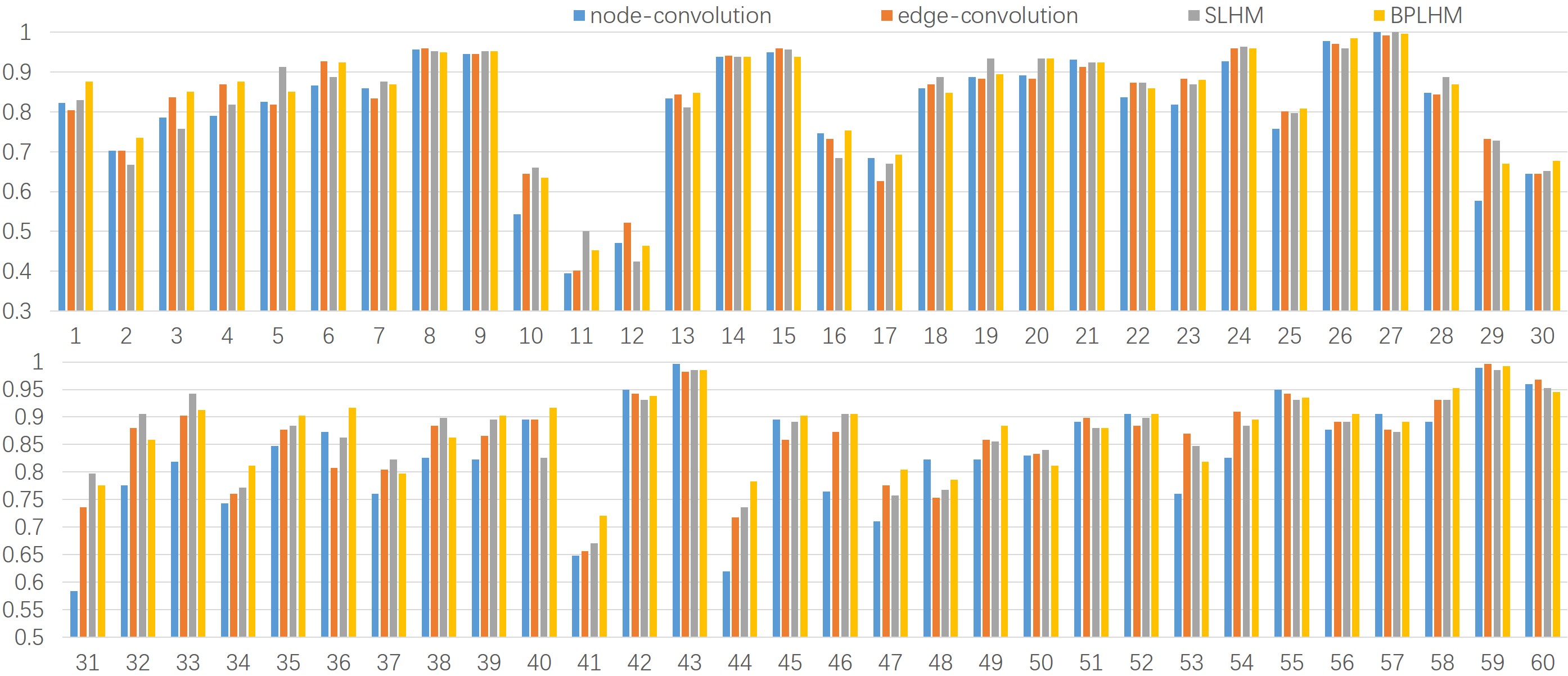}
\caption{Class accuracy on NTU-RGB+D dataset with cross-subject separation. For each class, we represent the accuracies of four models with bars in four different colors. We only provide the number denoting each class, for detailed name of each class, please refer to \href{https://github.com/shahroudy/NTURGB-D}{NTU-RGB+D}}
\label{fig:example}
\end{figure*}

{\bf Advantages of GECNN on certain classes:}
In Fig. 12, we report classification accuracy for each class, and if we have a detailed observation and consideration on certain classes in which edge convolution model greatly outperforms node convolution model, we can have more interesting findings and leads us to realize more advantages of recognizing human action from dynamics of bones. For example, in class 3, 10, 44, 47, 53, edge convolution model significantly outperforms node convolution model. And we can see that in these action classes, some joints are likely to overlap with other joints. e.g. in class 3, when brushing teeth, hand is overlapped with head (head is also represented as a joint point in skeletal data); in class 10, two hands overlap with each other when clapping; in class 44, hand overlaps with head when touching head; in class 47, hand overlap with neck when touching neck (neck is either represented as a joint point or ignored in skeletal data); in class 53, when patting on back of other person, the hand is likely to overlap with certain joint point located on the back. We suspect that overlapping between joint points may cause the lowering of accuracy of node convolution model. Because joints are relatively small compared to the whole environment in which actors conduct actions, thus if the distance between two joints is too small, the location of the joints and the dynamics of the joints will become difficult to capture (Fig. 11).

 However, things are different for bones, as bones are long parts that extend a certain distance in 3D space, it's difficult for two bones to totally overlap with each other. Two bones may cross at a point, but the major parts of the two bones are still separated. Thus, in these cases, although some of the joints are difficult to be distinguished from each other, the bones are always easy to be separated. Thus, besides the advantages of recognizing human action by bone dynamics we mentioned in Introduction section, the robustness against joint overlapping is also an advantage of utilizing dynamics of bones to do human action recognition.

In above paragraph, we analyze the advantages of edge convolution model by finding out in which classes it outperforms other models significantly and then analyze the characteristics of these classes. Similar analysis can also be conducted on other classes to find other advantages of different models, but we are not going to discuss further in this way, because this kind of analysis is not so rigorous and needs further validation to be fully convincing. Our discussion in the paragraph above is to provide a possible way to analyze the characteristics of different models, and more detailed validation is left for future work.

{\bf Model comparision on class accuracies:}
Besides the accuracy reported in Table 2, we can also compare the performance of the four models from another point of view. In Fig. 12, we can count that in 25 classes, our body-part level hybrid model outperforms other three models, in 18 classes, sequence level hybrid model performs the best, in 15 classes, edge convolution model gets the highest accuracies, and in only 8 classes, node convolution model gets better result than other three models. (In some classes, two of the three models may achieve same accuracy, thus the four numbers: 25,18,15,8 do not sum to 60, which is the total number of classes).
%这句话位置不对 
The result that in more classes GECNN gets higher accuracies than node convolution model also corresponds to our analysis in the introduction that in many cases, the movements of bones is more obvious and easier to capture. 
But we also note that on some classes, node convolution model gets the highest accuracies, demonstrating the indispensable role of joint dynamics, which is the reason we design hybrid models.

The hybrid models outperform the two models that based solely on joints or bones in both overall accuracy and number of best-performance classes, which demonstrates that our hybrid models successfully inherit the advantages from both bone based model and joint based model. 
Moreover, the result that body-part level hybrid model outperforms sequence level hybrid model tell us that combination of node and edge convolution will let the data utilized more throughly. Letting the information flow between connected joints and bones by involving them in the same convolution layer make joints and bones corroborate each other more accurately, while in sequence level hybrid model, the feature sets of bones and joints are separately extracted, and information exchange only happen between features of whole video sequences, which is rougher than body-part level hybrid model.

\section{Conclusion}
In this paper, we propose graph edge convolutional neural network, which is a novel way to conduct convolution operation on graph data. Our model captures the relationship and dependencies between edges by conducting convolution on edges, which is different from previous graph convolution methods that convolve nodes. Convolution on edges also enables us to leverage the dynamics of bones, and we apply our model to human action recognition tasks on two datasets, resulting in remarkably superior performance over existing state-of-the-art methods. 

Moreover, we also seek to combine node convolution and edge convolution together into two hybrid models that inherit advantages from both models, and the experiment result shows that our hybrid models can further improve the performance.

\bibliographystyle{IEEEtran}
\bibliography{egbib}

\end{document}